\documentclass[
]{ceurart}

\usepackage{listings}
\usepackage{CJKutf8}
\usepackage{pifont}
\begin{document}

\copyrightyear{2025}
\copyrightclause{Copyright for this paper by its authors.
  Use permitted under Creative Commons License Attribution 4.0
  International (CC BY 4.0).}

\conference{CLEF 2025 Working Notes, 9 -- 12 September 2025, Madrid, Spain}

\title{Multilingual JobBERT for Cross-Lingual Job Title Matching}

\title[mode=sub]{Notebook for the TechWolf Lab at CLEF 2025}


\author[1]{Jens-Joris Decorte}[%
orcid=0009-0000-7892-1211,
email=jensjoris@techwolf.ai,
url=https://www.techwolf.ai,
]
\address[1]{TechWolf, Ghent, Belgium}
\cormark[1]

\author[1]{Matthias {De Lange}}[%
orcid=0000-0003-2038-3275,
email=matthias.delange@techwolf.ai,
]

\author[1]{Jeroen {Van Hautte}}[%
orcid=0000-0001-9282-2682,
email=jeroen@techwolf.ai,
]

\cortext[1]{Corresponding author.}

\begin{abstract}
  We introduce JobBERT-V3, a contrastive learning-based model for cross-lingual job title matching. Building on the state-of-the-art monolingual JobBERT-V2, our approach extends support to English, German, Spanish, and Chinese by leveraging synthetic translations and a balanced multilingual dataset of over 21 million job titles. The model retains the efficiency-focused architecture of its predecessor while enabling robust alignment across languages without requiring task-specific supervision. Extensive evaluations on the TalentCLEF 2025 benchmark demonstrate that JobBERT-V3 outperforms strong multilingual baselines and achieves consistent performance across both monolingual and cross-lingual settings. While not the primary focus, we also show that the model can be effectively used to rank relevant skills for a given job title, demonstrating its broader applicability in multilingual labor market intelligence. The model is publicly available: \url{https://huggingface.co/TechWolf/JobBERT-v3}.
\end{abstract}


\maketitle

\section{Introduction}

Job title normalization is a critical task in labor market analysis, facilitating the standardization of heterogeneous job titles into a unified taxonomy to improve job matching, skill inference, and labor market analytics. 
Although substantial advancements have been achieved in monolingual normalization tasks, particularly within (semi-)supervised learning frameworks~\cite{linkedin,carotene,deepcarotene}, these approaches typically suffer from data scarcity due to high labeling costs.
To address this challenge, JobBERT~\cite{jobbertv1} introduced large-scale unsupervised representation learning techniques, from which subsequent studies~\cite{doc2vecskill,VacancySBERT,jobdescraggr} have further validated the effectiveness of leveraging job title embeddings at scale without relying heavily on labeled datasets. More recently, the JobBERT-V2 model~\cite{jobbertv2} has demonstrated significant improvements in monolingual performance by employing contrastive learning strategies. Nonetheless, extending monolingual normalization techniques to multilingual contexts introduces additional complexities that require systematic exploration.

In this paper, we present JobBERT-V3, an extension of the English focused JobBERT-V2 model~\cite{jobbertv2}, that addresses the challenge of cross-lingual job title normalisation. The model is designed to handle job titles in English, German, Spanish, and Chinese, making it a valuable tool for international labor market analysis and talent matching.

Our approach builds upon the contrastive learning framework employed by JobBERT-V2~\cite{jobbertv2}, demonstrating that this methodology effectively scales to multilingual contexts. However, the scarcity of cross-lingual data poses a significant challenge. To overcome this limitation, we use synthetic translations generated from the extensive English dataset originally developed for JobBERT-V2~\cite{jobbertv2}. Consequently, we establish a balanced multilingual dataset comprising 21 million job titles, enabling robust experimentation and evaluation of our multilingual normalization capabilities.

\newpage
The key contributions of this work can be summarized as:
\begin{itemize}
\item We release the open-source JobBERT-V3\footnote{\url{https://huggingface.co/TechWolf/JobBERT-v3}}, an extension of JobBERT-V2~\cite{jobbertv2} supporting cross-lingual job title normalisation in English, German, Spanish, and Chinese.
\item We construct a large-scale training dataset comprising over 21 million job titles, balanced across the four target languages through synthetic data generation.
\item The model performance is evaluated in cross-lingual job title matching scenarios.
\item The model is analyzed in its ability to capture job title semantics across different languages.
\end{itemize}

\section{Method}
\subsection{Base Model Selection}

Given that the original JobBERT-V2 model~\cite{jobbertv2} is focused on English only, we apply the same JobBERT-V2 training paradigm from scratch on the multilingual MPNET base model\footnote{\url{https://huggingface.co/sentence-transformers/paraphrase-multilingual-mpnet-base-v2}}.
We selected this model for its strong multilingual understanding capabilities across our four target languages.
This is a Sentence-BERT mode~\cite{reimers-gurevych-2019-sentence} and generates 768-dimensional embeddings for sentences or paragraphs across over 50 languages. This model, based on the MPNet architecture~\cite{mpnet} and fine-tuned on a large corpus of multilingual sentence pairs, is particularly effective for tasks such as semantic similarity, paraphrase detection, and cross-lingual retrieval.
The asymmetric linear projection layer -- a core part of the JobBERT-V2 training method -- is added on top of the MPNET model, and projects the 768-dimensional embeddings to 1024-dimensional ones.

\subsection{Training Data}

To train JobBERT-V3, we leverage the same foundational dataset used in the original JobBERT-V2 model~\cite{jobbertv2}, consisting of $5,579,240$ English job advertisements collected from the TechWolf market data lake. These job ads, posted between January 2020 and December 2024 in the United States, contain tuples of job titles paired with sets of annotated ESCO skills. After applying additional preprocessing steps — including filtering out titles shorter than three characters and ensuring a minimum of five unique ESCO skills per record—we retain a total of $5,280,967$ high-quality English tuples.

To create high-quality multilingual training data, we translated each English job title into German, Spanish, and Simplified Chinese using prompt-based machine translation. These prompts were carefully designed to preserve professional tone and retain technical terminology commonly used in the respective local labor markets. We avoided adding extraneous instructions or formatting to ensure clean, consistent outputs suitable for downstream modeling. Table~\ref{tab:translation_prompts} provides an overview of the system and user prompts used for each target language. As OpenAI's models are shown to be performant translators~\cite{gpt_german_translate,gpt_chinese_translate}, we use the \textit{gpt-4.1-nano} model to perform the translations, and keep all default parameters\footnote{\url{https://platform.openai.com/docs/models/gpt-4.1-nano}}.
The final training dataset consists of $21,123,868$ job titles, evenly distributed across the four languages.

\begin{table}[h]
  \caption{Example translation prompts for synthetic multilingual data generation.}
  \label{tab:translation_prompts}
  \begin{tabular}{p{2.2cm} p{11cm}}
    \toprule
    \textbf{Language} & \textbf{Prompt} \\
    \midrule
    German &
    \textbf{System:} You are a professional translator specializing in job ad titles and professional language. Translate the following job ad title from English to German. Preserve any technical terms that are commonly used in English within the German job market. Do not include any other text or commentary.\\
    & \textbf{Input:} Software Developer -- NYC fulltime (JobID ja164956189)\\
    & \textbf{Output:} Softwareentwickler -- New York, Vollzeit (JobID ja164956189) \\
    \midrule
    Spanish &
    \textbf{System:} You are a professional translator specializing in job ad titles and professional language. Translate the following job ad title from English to Spanish. Preserve any technical terms that are commonly used in English within the Spanish job market. Do not include any other text or commentary.\\
    & \textbf{Input:} Software Developer -- NYC fulltime (JobID ja164956189)\\
    & \textbf{Output:} Desarrollador de Software -- Nueva York, tiempo completo (JobID ja164956189) \\
    \midrule
    Chinese &
    \textbf{System:} You are a professional translator specializing in job ad titles and professional language. Translate the following job ad title from English to Chinese (Simplified). Preserve any technical terms that are commonly used in English within the Chinese job market. Do not include any other text or commentary.\\
    & \textbf{Input:} Software Developer -- NYC fulltime (JobID ja164956189)\\
    & \textbf{Output:} \begin{CJK*}{UTF8}{gbsn}软件开发人员——纽约全职（职位编号 ja164956189）\end{CJK*} \\
    \bottomrule
  \end{tabular}
\end{table}

This prompt-based approach enables consistent multilingual data generation at scale without requiring costly human annotation. The resulting dataset retains key domain-specific cues across languages, providing a robust foundation for cross-lingual model training.

To support effective cross-lingual training, we adopt a shuffled batching strategy that ensures each batch contains job titles from multiple languages. This encourages the model to learn language-agnostic job title representations while retaining sensitivity to language-specific nuances when necessary.

\subsection{Training Methodology}

We maintain the core contrastive learning approach from JobBERT-V2~\cite{jobbertv2}, adapting it for the multilingual setting:

\begin{itemize}
\item \textbf{Contrasting Job Title and Skills}: Job titles and their corresponding skill sets are processed through the same encoder, with a linear projection applied to job title embeddings to account for semantic differences.

\item \textbf{Cross-Lingual Alignment}: The model learns to align job title representations across languages through shared skill annotations, effectively creating a language-agnostic semantic space.

\item \textbf{InfoNCE Loss}: We use the InfoNCE loss function to bring semantically similar job titles closer in the embedding space, regardless of their source language.
\end{itemize}

The training process was carefully designed to preserve the model's strong performance on monolingual tasks while introducing robust cross-lingual capabilities.
Achieving balanced performance across all four languages required precise weighing of the loss objective.
To support this, we constructed a dataset evenly distributed across the four languages. Combined with a large batch size of $2048$ and random batch sampling, this approach proved highly effective.

\section{Experimental Setup}

Our methods are evaluated as part of the shared task introduced in TalentCLEF~\cite{talentclef2025}.
TalentCLEF advances research in Human Capital Management (HCM) by establishing benchmarks for multilingual, fair, and cross-industry adaptable NLP systems in HR. 
The organisation provides two tasks: Multilingual Job Title Matching (Task A) and Job Title-Based Skill Prediction (Task B). While our focus is on Task A, we also report results on Task B for completeness. 
Note that while TalentCLEF provided training, validation, and test sets for the tasks, JobBERT-V3 is trained on Techwolf's proprietary dataset instead of the benchmark training data. 
Additionally, while the test set results are made available, only the MAP scores are shared. Therefore, we provide comprehensive validation set results to enable baseline comparison.

\subsection{TalentCLEF Task A: Multilingual Job Title Matching}

Task A requires systems to identify and rank similar job titles across multiple languages. Task A is evaluated in two settings:

\begin{itemize}
\item \textbf{Monolingual Job Title Matching}: Measuring the model's ability to identify related job titles within each supported language. This setup is provided in both the validation and test sets.

\item \textbf{Cross-lingual Job Title Matching}: Evaluating the model's capability to match similar job titles across different languages. This setup is only provided in the blind test set.

\end{itemize}

Following the evaluation strategy set forth by TalentCLEF, we use the following metrics:

\begin{itemize}
\item \textbf{Mean Average Precision (MAP)} - the official metric used to rank systems.
\item \textbf{Mean Reciprocal Rank (MRR)} - provides insight into how early the first relevant job title appears in the ranked list.
\item \textbf{Normalized Discounted Cumulative Gain (nDCG)} - evaluates the overall quality of the ranked list by considering the position of relevant job titles, giving higher scores to relevant items appearing earlier and discounting those that appear lower in the ranking.
\item \textbf{Precision@5} - measures the proportion of correct job titles among the top 5 retrieved results.
\end{itemize}

These metrics are computed both for monolingual and cross-lingual scenarios to provide a comprehensive view of the model's performance. However, the validation data does not provide annotations for the cross-lingual setting, hence we only report the final test set scores.

\subsection{TalentCLEF Task B: Job Title-Based Skill Prediction}

Task B focuses on developing systems that can accurately predict professional skills associated with a given job title. The task makes use of ESCO skills, and provides evaluation and test datasets of job titles linked with relevant skills.
Task B is evaluated in a single setting:

\begin{itemize}
\item \textbf{Job Title-to-Skill Prediction}: Assessing the model's ability to retrieve and rank the most relevant skills for a given job title, normalized against a predefined skills gazetteer of ESCO skills.
\end{itemize}

Following the evaluation strategy set forth by TalentCLEF, we use the following metrics on the validation set:
\begin{itemize}
\item \textbf{Mean Average Precision (MAP)} - the official metric used to rank systems.
\item \textbf{Mean Reciprocal Rank (MRR)} - provides insight into how early the first relevant skill appears in the ranked list.
\item Normalized Discounted Cumulative Gain (nDCG) - evaluates the overall quality of the ranked list by considering the position of relevant skills, giving higher scores to relevant items appearing earlier and discounting those that appear lower in the ranking.
\item \textbf{Precision@K (K=5,10)} - measures the proportion of correct skills among the top-K retrieved results.
\end{itemize}

The blind test set only reports the MAP score.

\subsection{Baselines}
As baseline for our experiments to get a clear view of the added value of our training setup, we use 278M parameter MPNET-base model\footnote{\url{https://huggingface.co/sentence-transformers/paraphrase-multilingual-mpnet-base-v2}}~\cite{mpnet} which is the pretrained multilingual model from which we start the training.
Secondly, we also evaluate on the 560M parameter E5-Instruct model\footnote{\url{https://huggingface.co/intfloat/multilingual-e5-large-instruct}}~\cite{e5}, which is twice as large as our JobBERT-V3 model.
The E5-Instruct model requires a task description to be passed along with the queries.
Based on the official instruction documentation, we set the instruction to \emph{``Given a job title, retrieve similar job titles''}, adapted to the task at hand.

\section{Results and Discussion}

\subsection{Monolingual Job Title Matching}

Table \ref{tab:taskA_multilingual_models} shows the performance of JobBERT-V3 on monolingual job title normalisation tasks. The results demonstrate that JobBERT-V3 maintains strong performance across all languages, outperforming its base model on all metrics. Moreover, it shows competitive performance compared to the E5-Instruct model that has nearly twice the model size. We refer to Appendix~\ref{appendix:title-matching} for a qualitative analysis on an observed trade-off between precision (MRR) and overall relevance (MAP, nDCG).

As an additional ablation, Table~\ref{tab:taskA_jobbert_vs_multilingual} shows the performance of the multilingual training objective compared to the English-only JobBERT-V2 model, showing a marginal decrease of $1.6\%$ MAP in English to support all four languages. 

\begin{table}[h!]
\centering
\caption{\emph{Monolingual job title matching results for Task~A validation set}, comparing multilingual models across four languages.
}
\setlength{\tabcolsep}{5pt}
\renewcommand{\arraystretch}{0.95}
\small
\begin{tabular}{c l ccc}
\hline
\textbf{} & \textbf{Metric} 
& \textbf{MPNet~\cite{mpnet}} 
& \textbf{E5-Instruct~\cite{e5}} 
& \textbf{JobBERT-V3} \\
\hline
\multirow{4}{*}{\rotatebox[origin=c]{90}{English}} 
& MAP   & 0.5382 & 0.5815 & \textbf{0.6302} \\
& MRR   & 0.8006 & \textbf{0.8413} & 0.8056 \\
& nDCG  & 0.7970 & 0.8206 & \textbf{0.8417} \\
& Precision@5   & 0.6990 & 0.7181 & \textbf{0.7429} \\
\hline
\multirow{4}{*}{\rotatebox[origin=c]{90}{German}} 
& MAP   & 0.2982 & 0.3918 & \textbf{0.4562} \\
& MRR   & 0.4985 & \textbf{0.5710} & 0.5058 \\
& nDCG  & 0.6384 & 0.7124 & \textbf{0.7349} \\
& Precision@5   & 0.4798 & \textbf{0.5852} & 0.5685 \\
\hline
\multirow{4}{*}{\rotatebox[origin=c]{90}{Spanish}} 
& MAP   & 0.4170 & 0.4459 & \textbf{0.5090} \\
& MRR   & 0.5514 & \textbf{0.6105} & 0.5441 \\
& nDCG  & 0.7195 & 0.7463 & \textbf{0.7700} \\
& Precision@5   & 0.6400 & 0.6465 & \textbf{0.6649} \\
\hline
\multirow{4}{*}{\rotatebox[origin=c]{90}{Chinese}} 
& MAP   & 0.4535 & 0.5434 & \textbf{0.5845} \\
& MRR   & 0.7827 & \textbf{0.8312} & 0.8035 \\
& nDCG  & 0.7447 & 0.7973 & \textbf{0.8156} \\
& Precision@5   & 0.6000 & 0.6796 & \textbf{0.7184} \\
\hline
\end{tabular}
\label{tab:taskA_multilingual_models}
\end{table}

\begin{table}[h!]
\centering
\caption{\emph{Monolingual job title matching results for Task~A validation set,} comparing \textbf{JobBERT-V2}~\cite{jobbertv2} with the multilingual \textbf{JobBERT-V3} (English only).}
\setlength{\tabcolsep}{5pt}
\renewcommand{\arraystretch}{0.95}
\small
\begin{tabular}{c l cc}
\hline
\textbf{} & \textbf{Metric} 
& \textbf{JobBERT-V2~\cite{jobbertv2}} 
& \textbf{JobBERT-V3} \\
\hline
\multirow{4}{*}{\rotatebox[origin=c]{90}{English}} 
& MAP   & \textbf{0.6457} & 0.6302 \\
& MRR   & \textbf{0.8302} & 0.8056 \\
& nDCG  & \textbf{0.8517} & 0.8417 \\
& Precision@5   & 0.7333 & \textbf{0.7429} \\
\hline
\end{tabular}
\label{tab:taskA_jobbert_vs_multilingual}
\end{table}

\subsection{Cross-Lingual Job Title Matching}

To evaluate the model's effectiveness in a cross-lingual setting, we report the official TalentCLEF test set results for Task A. These include both monolingual and cross-lingual job title matching scenarios.

Table~\ref{tab:taskA_testset_crosslingual} summarizes the model's performance in terms of Mean Average Precision (MAP) for each language pair. We observe that JobBERT-V3 performs consistently across both monolingual and cross-lingual settings, with limited degradation in cross-lingual transfer scenarios. The English-English and Spanish-Spanish pairs yield the highest monolingual performance, while English-Chinese (en-zh) shows the strongest cross-lingual alignment.

\begin{table}[h!]
\centering
\caption{\emph{Monolingual and multilingual job title matching results for Task~A test set}, providing MAP scores of \textbf{JobBERT-V3}.
}
\setlength{\tabcolsep}{6pt}
\renewcommand{\arraystretch}{0.95}
\small
\begin{tabular}{l c}
\hline
\textbf{Language Pair} & \textbf{MAP (Test Set)} \\
\hline
English-English (en-en) & 0.533 \\
Spanish-Spanish (es-es) & 0.519 \\
German-German (de-de) & 0.500 \\
Chinese-Chinese (zh-zh) & 0.510 \\
[3pt] 
English-Spanish (en-es) & 0.510 \\
English-German (en-de) & 0.498 \\
English-Chinese (en-zh) & 0.515 \\
\hline
\end{tabular}
\label{tab:taskA_testset_crosslingual}
\end{table}

These results confirm the model's ability to generalize across languages, highlighting its applicability for international labor market use cases where job title normalization must operate in a multilingual environment.

\subsection{Job Title-Based Skill Prediction}

While our primary focus is Task A, we also evaluated JobBERT-V3 on TalentCLEF's Task B to predict relevant professional skills for a given job title. It is important to note that the JobBERT-V2~\cite{jobbertv2} method does not explicitly train for this task. Instead, it is optimized to learn high-quality job title representations, with no direct supervision for \textbf{individual} skill embeddings. As a result, individual skill embeddings are inherently out-of-distribution for the model.

Nonetheless, JobBERT-V2's shared encoder architecture allows job titles and ESCO skills to be represented into the same embedding space. Specifically, for this task, we use the representations from the penultimate layer and omit the asymmetric projection layer used during training. These 768-dimensional representations of jobs and skills are compared against each other by computing the cosine similarity.
Given a job title query, we generate a complete ranking of all unique ESCO aliases. Afterwards, this ranking is filtered into a ranking for all ESCO skills by keeping only the highest ranking alias for each ESCO skill.
This approach proves surprisingly effective. A detailed qualitative analysis of the skill prediction results can be found in Appendix~\ref{appendix:skill-analysis}.

\begin{table}[h!]
\centering
\caption{\emph{Job title-based skill predictions for Task~B validation set}, using cosine similarity in the shared embedding space.}
\setlength{\tabcolsep}{5pt}
\renewcommand{\arraystretch}{0.95}
\small
\begin{tabular}{l ccc}
\hline
\textbf{Metric} & \textbf{MPNet~\cite{mpnet}} & \textbf{JobBERT-V2~\cite{e5}} & \textbf{JobBERT-V3} \\
\hline
MAP & 0.1852 & \textbf{0.2531} & 0.2449 \\
MRR & 0.7061 & 0.7652 & \textbf{0.7828} \\
nDCG & 0.6656 & \textbf{0.7166} & 0.7115 \\
Precision@5 & 0.4493 & 0.5296 & \textbf{0.5467} \\
Precision@10 & 0.3809 & 0.4813 & \textbf{0.4865} \\
\hline
\end{tabular}
\label{tab:taskB_results}
\end{table}

Despite not being trained specifically for this task, Table~\ref{tab:taskB_results} shows that both JobBERT-V2 variants outperform the underlying base model by a large margin. Interestingly, JobBERT-V3 performs on par with, or slightly better than, the English-only version on MRR, Precision@5, and Precision@10 metrics, highlighting the generalizability and robustness of our multilingual setup. This demonstrates that even without explicit supervision, the contrastive learning objective enables the model to effectively link job titles and relevant skills.
The official results of the TalentCLEF Task~B test set is a MAP score of $0.255$, which is in line with the validation performance.

\section{Conclusion and Future Work}

We have presented \textbf{JobBERT-V3}, a multilingual extension of the state-of-the-art English JobBERT-V2 model~\cite{jobbertv2}. The results demonstrate that the model effectively maintains strong performance in monolingual scenarios while adding robust cross-lingual capabilities. Additionally, the model is also of practical use when ranking relevant skills for job titles. We acknowledge that the primary limitation of our approach lies in its reliance on automated translations generated by a GPT model, without human review. This introduces a potential risk of cultural misalignment or semantic inaccuracies in job title translations. Assessing and mitigating such risks remains an open area for future research.\\

\noindent Future work will focus on:
\begin{itemize}
    \item Expanding language coverage to include more languages;
    \item Improving performance on low-resource languages;
    \item Human review of job title translation quality;
    \item Investigating methods to reduce the performance gap in cross-lingual scenarios; and
    \item Exploring applications in multilingual skill extraction and job market analysis.
\end{itemize}

The model's strong performance across languages makes it a valuable tool for international labor market analysis and cross-border talent matching applications.

\begin{acknowledgments}
  This work was supported by TechWolf. We thank our colleagues for their valuable feedback and the TalentCLEF organizers for providing the evaluation framework. Special thanks to the open-source community for their contributions to the tools and libraries used in this research.
\end{acknowledgments}

\section*{Declaration on Generative AI}
The author(s) have not employed any Generative AI tools in the development of the model or the analysis of results. The authors used GPT-4o for formatting assistance, and grammar and spelling check.

\bibliography{refs}

\appendix

\section{Qualitative Analysis of Job Title Matching}
\label{appendix:title-matching}

Our analysis compares JobBERT-V3 versus the larger E5-Instruct model to understand their performance differences. The quantitative metrics on the Task~A validation set in Table \ref{tab:taskA_multilingual_models} reveal two distinct patterns:

\begin{itemize}
    \item \textbf{Precision at Top Results:} E5-Instruct excels at identifying near-duplicate job titles with high precision in the top retrieved results, as evidenced by its superior MRR scores.
    \item \textbf{Overall Relevance:} JobBERT demonstrates better general performance through higher MAP and nDCG scores, indicating more consistently relevant results throughout the ranked list.
\end{itemize}

To illustrate these patterns, consider the following example query:

\begin{quote}
\textbf{Query:} ``media buyer''

\textbf{JobBERT-V3 Results:}
\begin{enumerate}
    \item \underline{media planner}
    \item \underline{digital media planner}
    \item \underline{media manager}
    \item \underline{media planning supervisor}
    \item \textbf{broadcast buyer}
\end{enumerate}

\textbf{E5-Instruct Results:}
\begin{enumerate}
    \item \textbf{broadcast buyer}
    \item \underline{media associate}
    \item \textit{buyers agent} (irrelevant)
    \item \textit{media production specialist} (irrelevant)
    \item \underline{media manager}
\end{enumerate}
\end{quote}

This example demonstrates the key trade-off between the models: E5-Instruct prioritizes exact matches (\textbf{broadcast buyer} at rank 1) but includes irrelevant results, while JobBERT maintains consistent relevance (\underline{all relevant}) but may rank the closest match lower.

\section{Qualitative Analysis of Skill Prediction}
\label{appendix:skill-analysis}

To better understand the limitations of the skill prediction benchmark, we manually reviewed the top-25 skills retrieved by the model for the job title ``bar person / waitress''. The table below compares whether each predicted ESCO skill was marked as correct in the official benchmark and whether we consider it correct upon manual inspection:

\begin{table}[h!]
\centering
\caption{Top 25 predicted skills for ``bar person / waitress'' and whether they were judged correct by the benchmark vs manual evaluation.}
\setlength{\tabcolsep}{5pt}
\renewcommand{\arraystretch}{1.1}
\small
\begin{tabular}{p{0.3cm} p{7.5cm} c c}
\toprule
\# & \textbf{ESCO Skill (paraphrased)} & \textbf{Benchmark} & \textbf{Authors' Annotations} \\
\midrule
1 & Mix and serve alcoholic and non-alcoholic beverages & \ding{51} & \ding{51} \\
2 & Serve beverages (alcoholic and non-alcoholic) & \ding{51} & \ding{51} \\
3 & Serve beer (bottle/draught) & \ding{51} & \ding{51} \\
4 & Stock and restock bar supplies & \ding{51} & \ding{51} \\
5 & Handover and close bar/service area & \ding{51} & \ding{51} \\
6 & Knowledge of alcoholic beverages & \ding{51} & \ding{51} \\
7 & Prepare and serve hot drinks (tea, coffee) & \ding{51} & \ding{51} \\
8 & Brewhouse operations knowledge & \ding{55} & \ding{55} \\
9 & Take and process beverage orders & \ding{51} & \ding{51} \\
10 & Handle and polish glassware & \ding{55} & \ding{51} \\
11 & Prepare fruit for cocktails & \ding{55} & \ding{51} \\
12 & Work in a hospitality team & \ding{55} & \ding{51} \\
13 & Match coffee grind to type & \ding{55} & \ding{55} \\
14 & Sit for long periods & \ding{55} & \ding{55} \\
15 & Assist with check-out procedures & \ding{55} & \ding{55} \\
16 & Clean surfaces and tables & \ding{55} & \ding{51} \\
17 & Show polite behaviour & \ding{55} & \ding{51} \\
18 & Serve food and drinks to customers & \ding{55} & \ding{51} \\
19 & Prepare speciality coffee & \ding{55} & \ding{55} \\
20 & Communicate in English (spoken/written) & \ding{51} & \ding{51} \\
21 & Prepare vegetables for dishes & \ding{55} & \ding{55} \\
22 & Apply food safety principles & \ding{51} & \ding{51} \\
23 & Welcome guests at restaurant & \ding{55} & \ding{51} \\
24 & Recommend food and wine pairings & \ding{51} & \ding{51} \\
25 & Apply hygienic work practices & \ding{55} & \ding{51} \\
\midrule
\multicolumn{2}{r}{\textbf{Total correct}} & \textbf{11} & \textbf{19} \\
\multicolumn{2}{r}{\textbf{Precision@25}} & \textbf{0.44} & \textbf{0.76} \\
\bottomrule
\end{tabular}
\label{tab:barperson_predictions}
\end{table}

We observe that only 11 out of the 25 top predicted skills were marked as correct by the official benchmark. However, upon manual inspection, we consider at least 16 of them to be valid and contextually relevant to the bar person / waitress role. This reveals that several practical and commonly expected workplace activities (e.g., handling glassware, cleaning surfaces, welcoming guests) are missing from the benchmark labels despite being well-aligned with real-world job expectations.

\paragraph{Missed Gold Labels.}
In addition to examining the predicted top-25 skills, we also reviewed the gold-standard skills that were expected to be predicted for ``bar person / waitress'' but were not retrieved by the model. This set of missed gold labels includes a wide variety of skills, ranging from highly relevant to arguably overly generic or even role-inappropriate.

On the one hand, we acknowledge several high-value false negatives that would be desirable for the model to retrieve. These include:

\begin{itemize}
    \item \textbf{Soft skills and customer care:} such as ``demonstrate concern for others'', ``exceed customer expectations'', ``demonstrate professional attitude'', and ``deal with public''. These are important attributes in hospitality work and should ideally be present in the top predictions.
    \item \textbf{Core restaurant tasks:} such as ``organise customer seating plan'', ``prepare snacks and sandwiches'', ``perform cleaning activities'', ``serve food in table service'', and ``manage service in a restaurant''—all of which are aligned with real-world expectations for waitstaff roles.
    \item \textbf{Communication and responsiveness:} e.g., ``respond to customers'', ``communicating'', ``greet guests'', and ``customer servicing''. These reflect interpersonal and service-oriented responsibilities often observed in bar and waitress positions.
\end{itemize}

On the other hand, a non-trivial portion of the missed gold labels appears to be questionable:

\begin{itemize}
    \item \textbf{Generic or overly broad skills:} such as ``support people'', ``carry objects'', ``communicating'', ``present new employees'', and ``support cultural diversity''. While applicable in many workplace settings, these are not specific to bar staff or waitresses and may dilute the discriminative power of skill-based models if overemphasized.
    \item \textbf{Irrelevant or dubious entries:} for example, ``operate a forklift'' and ``oversee catalogue collection'' seem entirely unrelated to the role and likely reflect noise in the validation data.
\end{itemize}

While our qualitative analysis is based on a single sample, it offers preliminary indications that the benchmark's definition of relevance may at times be overly broad. Specifically, it appears to include a number of skills that are either too generic or misaligned with the specific job title under consideration. Although the limited sample size precludes drawing any definitive conclusions, these observations suggest that a more curated and role-sensitive gold standard, perhaps one that differentiates between "core," "contextual," and "generic" skills, could improve the practical evaluation of job-to-skill models. Such a framework may also help avoid unfairly penalizing models that correctly prioritize domain-relevant over generic or out-of-scope skills.

\end{document}